%% file: root.tex
\documentclass[runningheads]{llncs}
\RequirePackage{silence}  
\usepackage{graphicx}
\usepackage{comment}
\usepackage{amsmath,amssymb}
\usepackage{color}
\usepackage{url}
\usepackage{hyperref}

\usepackage{dsfont}

\usepackage{pifont}
\usepackage{tabularx}
\usepackage{makecell}
\usepackage{multirow}
\usepackage{booktabs}
\newcolumntype{C}{>{\centering\arraybackslash}X}
\newcolumntype{L}{>{\raggedright\arraybackslash}X}
   
\newif\ifreview
\reviewtrue
\reviewfalse

\ifreview
	\usepackage{lineno}

	\linenumbers
\fi

\begin{document}

\title{ARETE: Attention-based Rasterized Encoding for Topology Estimation using HSV-transformed Crowdsourced Vehicle Fleet Data}

\ifreview
\else
	\titlerunning{Topology Estimation using Crowdsourced Vehicle Fleet Data}

	\author{Daniel Fritz\inst{1,2} \and
	Dimitrios Lagamtzis\inst{1} \and
	Michael Mink\inst{1} \and Markus Enzweiler\inst{2} \and Steffen Schober\inst{2}}
	
	\authorrunning{D. Fritz et al.}

	\institute{Mercedes-Benz AG, Research \& Development, Sindelfingen, Germany \and Institute for Intelligent Systems, Esslingen University of Applied Sciences, Esslingen, Germany}
\fi

\maketitle              

\begin{abstract}
The continuous advancement of autonomous driving (AD) introduces challenges across multiple disciplines to ensure safe and efficient driving. One such challenge is the generation of High-Definition (HD) maps, which must remain up to date and highly accurate for downstream automotive tasks. One promising approach is the use of crowdsourced data from a vehicle fleet, representing road topology and lane-level features. This work focuses on the generation of centerlines and lane dividers from crowdsourced vehicle trajectories. We adopt a Detection Transformer (DETR)-based approach, where a rasterized representation of vehicle trajectories is used as input to predict vectorized lane representations. Each lane consists of a centerline with an associated direction and corresponding lane dividers that are geometrically constrained by the centerline. Our method includes the extraction of local tiles, from which crowdsourced vehicle trajectories are aggregated. Each tile undergoes a transformation into a rasterized representation encoding both the presence and direction of each trajectory, enabling the prediction of vectorized directed lanes. Experiments are conducted on an internal dataset as well as on the public datasets nuScenes and nuPlan.

\end{abstract}

\input{content/introduction}
\input{content/related_work}

\input{content/method}
\input{content/experiments}
\input{content/conclusion}
\input{content/acknowledgment}

\bibliographystyle{splncs04}
\bibliography{bib}

\end{document}

%% file: content/introduction.tex
\section{INTRODUCTION}
Digital maps are widely used by human drivers for navigation and spatial orientation when planning and reaching a desired destination. In this context, Standard-Definition (SD) maps provide a coarse representation of the road network, which captures fundamental geographic information that is generally sufficient for basic navigation tasks. When considering the task of Autonomous Driving (AD), the requirements of digital maps become significant stricter in terms of coverage and accuracy in order to represent the road structure and its lane-level features. A higher level of detail enables a more comprehensive understanding of the driving environment and supports downstream tasks such as motion planning, scene understanding, and localization \cite{elghazaly_map_survey}. 

Recent work focuses on generating centimeter-accurate HD maps by leveraging data from onboard sensors, such as multi-view camera images and point clouds from radar and LiDAR, to predict HD map elements within online systems \cite{lyu_map_generation_survey}. In contrast, offline HD map generation methods leverage crowdsourced data, such as vehicle trajectories, and can also incorporate additional data sources, such as SD maps or satellite imagery, to construct HD map elements \cite{guo_map_generation_survey}.


This work introduces a pipeline for generating vectorized lane representations consisting of a centerline and corresponding left and right lane dividers. A global city-level region is partitioned into local tiles, where crowdsourced ego- and tracked-vehicle trajectories are aggregated. These trajectories are transformed into a rasterized representation compatible with DETR-based architectures, encoding direction and presence via the HSV color model. To further enrich the representation, we incorporate mean velocity and spatial offsets lost during rasterization. The contributions of this paper are as follows:

\begin{itemize}
\item A pipeline for local tile extraction and vehicle trajectory aggregation.
\item A detection transformer, which takes as input a rasterized version of crowdsourced trajectories and generates vectorized lanes, explicitly modeling the relationship between a centerline and its corresponding lane dividers.
\item A comprehensive augmentation pipeline and extensive experiments on dataset combinations to improve generalization.
\end{itemize}

%% file: content/related_work.tex
\section{RELATED WORK}
\subsection{Detection Transformer}
DETRs \cite{detr} have gained popularity in object detection due to the omission of handcrafted features. This enables an end-to-end approach using learnable queries, each representing a potential object within the scene. This methodology has also been applied to online and crowdsourced HD map generation, leading to specialized approaches that leverage hierarchical learnable queries to encode instance- and point-level representations of HD map elements such as centerlines, lane dividers, road boundaries, and pedestrian crossings \cite{local_hd_gen_survey}.

\subsection{Online-based Map Generation}
Data from onboard sensors, such as cameras and LiDAR, serve as input for detecting and generating vectorized HD map elements, ultimately providing a strong prior for downstream tasks in automotive driving. In addition to high accuracy, real-time capability is essential for supplying geometric and semantic information about the vehicle's surroundings. For online HD map generation, camera-based data is most commonly used, from which bird’s-eye view (BEV) features are extracted and potentially fused with other data domains such as LiDAR point clouds. The resulting features are then fed into a transformer architecture to predict vectorized polylines, including centerlines, lane dividers, road boundaries, and pedestrian crossings, in an end-to-end approach \cite{vectormapnet,maptrv,streammapnet}. 

LaneGAP \cite{lane_gap} learns hierarchical queries to predict centerlines from BEV features. To model the connectivity between distinct centerlines, a post-processing step converts the individual predicted centerlines into a graph-based representation, enabling the formation of merging and branching structures. In contrast, LaneSegNet \cite{lanesegnet} directly models polylines at the segment level along with their connectivity. The framework employs a modified deformable attention mechanism to more effectively capture segment geometry. Lane dividers are predicted as offsets from embedded queries and subsequently added to the corresponding centerline segments to obtain the final segment-level lane representation. MapTRv2 \cite{maptrv2} introduced decoupled attention across hierarchical queries, and utilized a one-to-many assignment strategy between predictions and ground truth in order to accelerate convergence. The work in \cite{Hao_25} employs separate branches to extract features from camera and LiDAR data, which are subsequently processed by a transformer for cross-attention and fusion. Combined with data augmentation strategies that explicitly simulate sensor failures, this approach improves robustness and generalization. The model in \cite{bai_2026} uses BEV features from multi-view camera images to generate HD map elements. The queries are initially enhanced using BEV features, providing a prior before refinement within the Transformer module. A dynamic query management strategy is introduced to filter out queries with low confidence scores.

\subsection{Crowdsourced-based Map Generation}
Methods for offline HD map generation have the advantage of not being constrained by real-time requirements. Furthermore, they can leverage crowdsourced data, which allows for a more comprehensive representation of a given scene when generating such maps.

The framework proposed in \cite{Qin_23} identifies complex urban intersections and generates a vectorized representation based on crowdsourced data, including camera images and vehicle trajectories. For a given intersection, these two data modalities are aligned within a global coordinate system using a geometric transformer. By applying heuristic rules, the road topology of complex intersections is inferred, from which the most human-like paths are selected. However, careful tuning of the associated hyperparameters may be required when transferring the method to different datasets, particularly those with lower data density. In contrast, MapCVV \cite{mapcvv} leverages crowdsourced images to extract lane-level features, such as lane markings, stop lines, poles, traffic lights, and signs, in combination with crowdsourced trajectory data. The pipeline consists of multiple modules designed to reduce observational redundancy across different viewpoints and measurement runs within a scene, ultimately producing a vectorized HD map. While the aforementioned crowdsourced methods rely on rasterized representations, LMT-Net \cite{lmtnet} operates solely on vectorized inputs, including lane dividers and vehicle trajectories. Polylines are encoded as embedded one-dimensional vectors and processed by a Transformer using precomputed center points as queries and the encoded polylines as keys and values. The output consists of spatial point pairs and their connectivity, forming vectorized lanes as a directed graph. Experiments are conducted on an internal dataset, which limits reproducibility. The approach of TrailTR \cite{TrailTR} utilizes crowdsourced trajectories of both the ego vehicle and tracked agents. The vectorized data is transformed into a rasterized representation to enable compatibility with DETR-like architectures. The resulting images are split into multiple channels, where each channel is aggregated based on trajectory segment directions. This formulation enables the prediction of polylines with a single possible point-sequence permutation. Data augmentation is applied to improve generalization, however, the predictions are limited to centerlines. \cite{Lie_2026} uses the HSV color model to rasterize vehicle trajectories with directional information, along with average velocity profiles and corresponding satellite imagery of a local region. The model architecture incorporates adjacent regions to further improve accuracy. The work in \cite{fritz_advntg_2025} presents a pipeline to reduce the volume of crowdsourced data by applying simplification methods that remove redundancy while preserving the dataset’s characteristics, before passing the trajectories to a Transformer architecture for further polyline refinement.

While related work provides frameworks for HD map construction, none of them specifically investigate lane representation prediction where lane dividers are directly constrained by the centerline using solely crowdsourced trajectories as input. In this work, we aim to address this gap by conducting comprehensive experiments on datasets with different distributions of trajectory density, in order not only to predict the centerline of a lane they represent, but also the corresponding lane dividers, ultimately forming a lane.


%% file: content/method.tex
\section{METHOD}
This section provides the pipeline for processing the dataset consisting of vehicle trajectories. Given a city-level location, local tiles are extracted, where trajectories are aggregated along with the corresponding ground truth information. Subsequently, each tile is converted into a rasterized representation, serving as the final model input.

\begin{figure*}[!ht]
    \centering
    \includegraphics[width=\textwidth]{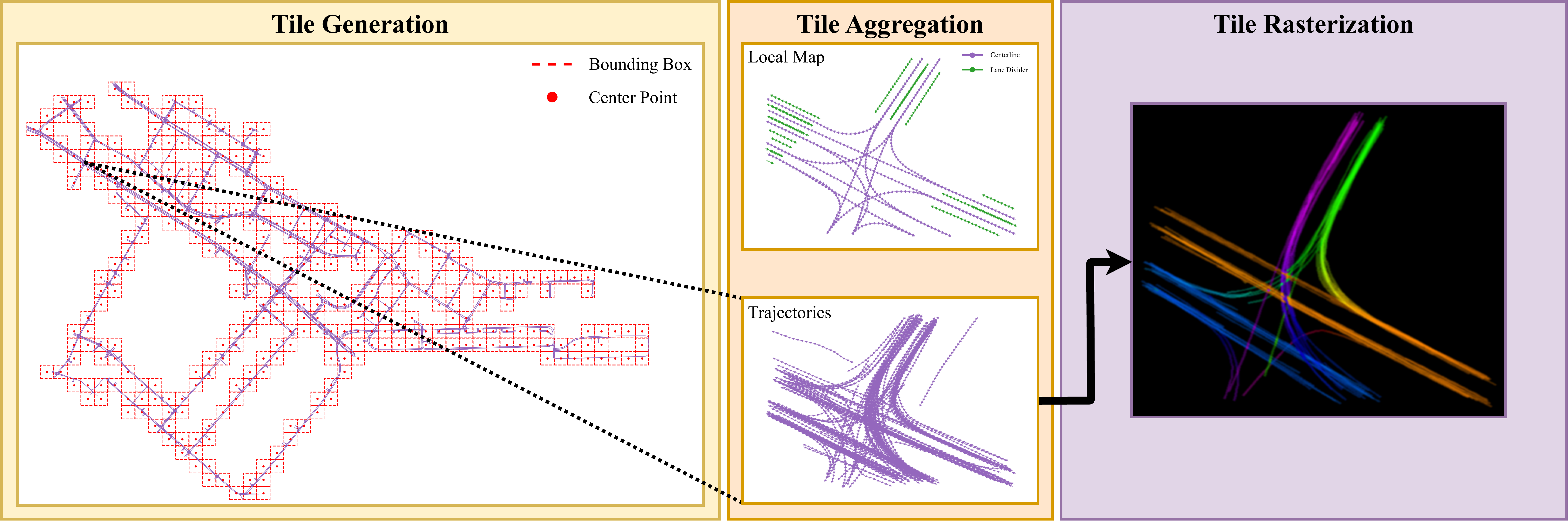} 
    \caption{Simplified illustration of the pipeline for generating local tiles. Tile selection is performed based on a location-level vectorized map. For each local tile, all possible ground truth paths are extracted, and corresponding vehicle trajectories are aggregated. Finally, each tile is rasterized using the aggregated vehicle trajectories.}
    \label{fig:tile_selection}
\end{figure*}

\subsection{Tile Generation}
\label{Tile Generation}
In \autoref{fig:tile_selection}, an example is shown of an vectorized map describing the road structure of a certain location. The map is modeled as a graph $\mathcal{G} = (\mathcal{V}, \mathcal{E})$, where $\mathcal{V}$ denotes the set of nodes (spatial points) and $\mathcal{E}$ represents the set of edges connecting these nodes.

We adopt a simple tile generation strategy by projecting a regular grid onto the map, where each resulting cell corresponds to a tile. More formally, a tile $t_i \in \mathcal{T}_j$ is defined as $t_i = (p_i, h_i, w_i)$, where $p_i \in \mathbb{R}^2$ denotes the global center point of the tile, and $h_i$ and $w_i$ denote its height and width, respectively. Here, $i$ indexes the $i$-th tile within location $j$, and $\mathcal{T}_j$ denotes the set of all tiles for the $j$-th location. The trajectory points of each individual tile are transformed into a local coordinate system, with the tile’s center serving as the origin. Since the requirement for a valid tile is the availability of both crowdsourced data and ground truth information, tiles that are missing either of the two are discarded.

It is worth noting that tile generation is not restricted to the grid-based method, which produces non-overlapping tiles. To increase data diversity, additional tiles are sampled around each grid-based tile, resulting in overlapping regions. However, this augmentation strategy is applied exclusively to the training split.

\subsection{Tile Aggregation}
\label{Tile Aggregation}
Given the tiles, data aggregation is performed to obtain a set of vehicle trajectories for each tile. Specifically, for a tile $t_i$, we collect the set of ego-vehicle trajectories $D_i$ and the set of trajectories from other tracked vehicles $O_i$. Combining these two sets yields a set that can be abstractly referred to as a set of polylines. Each polyline is represented by a temporal sequence of spatial points along with a velocity for each point.

The local ground truth information includes the local ground truth map representing the local region of the tile. Following LaneGAP \cite{graph_as_path}, we extract all possible paths from the local ground truth graph. Specifically, for each start node, paths are generated that terminate at distinct end nodes.

For our purpose, ground truth annotations for lane dividers are necessary, which are lacking particularly in the nuScenes dataset. In order to provide this information, we simulate lane dividers by exploiting the polygons encompassing the ground truth centerlines, which describe the drivable area of a lane segment. Hence, we perform lateral sampling along the centerline to create the lane divider information (which can also be seen as the boundaries of a lane). However, in certain scenes, such as intersections, these polygons are not available, leading to gaps when extracting lane dividers (see \autoref{fig:tile_selection}). Therefore, we perform interpolation to create complete vectorized lane representations, where for each center point there exist left and right lateral lane divider points.

Before transforming the data into a rasterized representation, vehicle trajectories that do not spatially align with the ground truth are removed, and the same filtering is applied to the ground truth polylines. This ensures that the crowdsourced and ground truth data are spatially consistent. We also prune the polylines such that the start and end points are approximately aligned with each other to enable fair evaluation.

\subsection{Tile Rasterization}
\label{Tile Rasterization}
To enable the prediction of polylines with directional information, the rasterized image must encode trajectory direction. Inspired by \cite{djuric_2020}, we employ the HSV color model, which can simultaneously represent both the presence of a trajectory and its direction. The HSV model consists of three components: \textbf{H}ue, \textbf{S}aturation, and \textbf{V}alue. The hue encodes the trajectory direction, while the value represents the brightness of the color, which we exploit to indicate the intensity, i.e., how frequently a pixel is occupied by points from vehicle trajectories. The saturation is set to 1. The HSV representation is then converted into an RGB image, where each RGB value corresponds to a direction angle, as illustrated in \autoref{fig:tile_selection}.

Furthermore, we extend the number of channels by including the pixel-wise mean velocity, as well as the x- and y-offsets for each occupied pixel, in order to preserve geometric information lost when transforming spatial points into discrete pixel coordinates. The final input representation is an image denoted as $x_{img} \in \mathbb{R}^{C \times H \times W}$, where $C$, $H$, and $W$ denote the number of channels, height, and width, respectively. In this case, the input has six channels: three from the RGB image converted from HSV, one for the velocity profile, and two for the x- and y-offsets.

%% file: content/experiments.tex
\section{EXPERIMENTS}
\subsection{Datasets}
\label{datasets}

\begin{table}[b!]
    \centering
    \caption{Statistics of trajectories and ground truth lanes per tile.}
    \label{tab:stats}
    \fontsize{9}{11}\selectfont
    \begin{tabularx}{\linewidth}{l *{4}{>{\centering\arraybackslash}X}}
        \toprule
        & \multicolumn{2}{c}{\textbf{Vehicle Trajectories \#}} & \multicolumn{2}{c}{\textbf{GT Lanes \#}} \\
        \cmidrule(lr){2-3} \cmidrule(lr){4-5}
        \textbf{Dataset} & Mean & Max & Mean & Max \\
        \midrule
        \midrule
        Internal  & 16 & 200 & 3 & 15 \\
        nuPlan    & 1500 & 94000 & 6 & 28 \\
        nuScenes  & 20 & 260  & 3 & 19 \\
        \bottomrule
    \end{tabularx}
\end{table}

For our experiments, we use the nuScenes dataset. It contains approximately 15 hours of driving data collected in Boston and Singapore, comprising 1,000 manually selected scenes, with each scene covering a time interval of 20 seconds. Following \cite{nuscenes_dataset_split}, the nuScenes dataset is split using a near-extrapolation split to prevent data leakage. Furthermore, we utilize nuPlan as a second public dataset, providing significantly more driving time (1,200 hours). The original splits are used for nuPlan. Training is also conducted on an internal dataset covering the German cities Sindelfingen and Boeblingen. Similar to nuScenes and nuPlan, our dataset provides trajectories of ego-vehicles and tracked vehicles. The internal dataset provides ground truth information about centerlines and lane dividers, which omits the need to interpolate lane divider information.

\autoref{tab:stats} summarizes the mean and maximum number of vehicle trajectories used for rasterization, as well as the number of ground truth lanes per tile. While the internal dataset and nuScenes exhibit similar distributions, nuPlan provides significantly denser data.

\subsection{Data Augmentation}
To increase data diversity, data augmentation is applied during training. As mentioned in Section \ref{Tile Generation}, the number of grid-based tiles is extended by randomly sampling tiles to expose the model to more diverse local regions. Furthermore, we apply various methods to modify the polylines of a tile before transforming it into an image, including random rotations and horizontal and vertical flips. To simulate different trajectory densities within a tile, sets of trajectories are randomly dropped, which ultimately leads to a different number of lanes to predict within a tile. Additionally, Gaussian noise is added to the x- and y-coordinates. Moreover, points are randomly shifted by sampling from a uniform distribution in the range $[-3.0, +3.0]$ along both axes. Finally, image masking is applied by randomly zeroing out patches within the input image. Each augmentation method is independently applied with a probability of 0.3, allowing multiple augmentations to be applied to a single tile.

\subsection{Model Architecture}

\begin{figure*}[!tb]
    \centering
    \includegraphics[width=\textwidth]{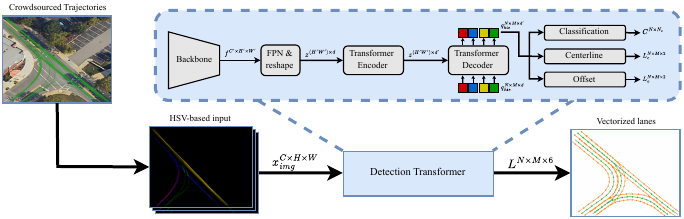} 
    \caption{Overview of the model for vectorized lane prediction. The input is a rasterized version of the crowdsourced vehicle trajectories. The deformable DETR processes the features from the backbone and outputs lane representations consisting of a centerline and its corresponding left and right lane dividers.}
    \label{fig:model}
\end{figure*}

As illustrated in \autoref{fig:model}, the input image $x_{img} \in \mathbb{R}^{C \times H \times W}$ is first passed through the backbone network, where $C$, $H$, and $W$ are the number of channels, height, and width, respectively. The resulting feature map $f \in \mathbb{R}^{C^{\prime} \times H^{\prime} \times W^{\prime}}$ is then processed by a Feature Pyramid Network (FPN) to transform the channel dimension $C^{\prime}$ into $d$. The dimensions $H^{\prime}$ and $W^{\prime}$ are reshaped into a single sequence, yielding a representation $z \in \mathbb{R}^{(H^{\prime}W^{\prime}) \times d}$. Subsequently, $z$ is processed by a deformable transformer encoder and decoder. As described in \cite{maptrv2}, decoupled attention over hierarchical queries $q^{hie} \in \mathbb{R}^{N \times M \times d}$ is applied, where $N$ and $M$ denote the number of queries and the number of points per query, respectively.

Finally, two regression branches are employed to generate vectorized lane from the learnable queries. The centerline branch predicts a sequence of points, denoted as $L_c \in \mathbb{R}^{N \times M \times 2}$, where $N$ represents the number of centerlines in a scene and $M$ denotes the number of points per centerline. Each point is defined by its Cartesian $x$- and $y$-coordinates in the local coordinate system of the tile. Inspired by LaneSegNet \cite{lanesegnet}, we further employ an offset branch, which takes the hierarchical queries as input and outputs the pointwise offsets for the $x$- and $y$-values, denoted as $L_o \in \mathbb{R}^{N \times M \times 2}$. Consequently, the left and right lane dividers can be computed as $L_l = L_c + L_o$ and $L_r = L_c - L_o$, respectively. The final lane can be denoted as $L = \{L_c, L_l, L_r\}$, with $L \in \mathbb{R}^{N \times M \times 6}$, since for each center point there exists a corresponding left and right lane divider point, each consisting of $x$- and $y$-values. The classification branch predicts the probability of each instance being either an object or a no-object, resulting in $N_c = 2$ classes.

\subsection{Training Loss}
We follow the methodology of DETR to compute the loss by performing bipartite matching between ground truth and predictions \cite{detr}. Specifically, we consider the sets of ground truth and predictions, $y=\{y\}_{i=1}^N$, $\hat{y}=\{\hat{y}\}_{i=1}^N$, respectively. To match the number of predictions, the ground truth set is padded with $\varnothing$ instances, representing the no-object class. The optimal assignment is then formulated as

\begin{equation}
    \hat{\sigma} = \underset{\sigma \in \mathfrak{S}^{N}}{\arg\min} \sum_{j}^{N} \mathcal{L}_{\text{match}}(\hat{y}_{j}, y_{\sigma_{(j)}}),
    \label{eq_1}
\end{equation}

\noindent where $\mathfrak{S}^{N}$ denotes the set of possible assignment permutations, and $\mathcal{L}_{\text{match}}$ represents the matching cost, consisting of the Manhattan distance for the geometric cost and the focal loss for the classification cost. Given the assigned pairs described by $\hat{\sigma}$, the classification loss is calculated as the focal loss

\begin{equation}
    \mathcal{L}_{\text{class}} = \sum_{i}^{N} \mathcal{L}_{\text{Focal}}(\hat{p}_{\hat{\sigma}_{(i)}}, c_i),
\end{equation}

\noindent where $\hat{p}_{\hat{\sigma}(i)}$ denotes the predicted classification score and $c_i$ the corresponding class label. The use of focal loss is motivated by the class imbalance, as the number of predictions is significantly higher than the average number of ground truth instances within a scene. The classification task is binary and serves solely to distinguish whether a prediction corresponds to a lane or to the no-lane (no-object) class. For the displacement loss, only the $N_v$ positively assigned pairs are considered. This loss is based on the Manhattan distance $D_{\text{L1}}$, and is formulated as

\begin{equation}
    \mathcal{L}_{\text{point}} = \sum_{i}^{N} \mathds{1}_{\{ c_i \neq \varnothing \}} \sum_{j}^{N_v} D_{L1} (\hat{y}_{\hat{\sigma}_{(i)},j}, y_{i, \sigma_{(j)}}),
\end{equation}

\noindent where the Manhattan distance is calculated pointwise over the centerline and lane divider points. To encourage the prediction of coherent centerlines and lane dividers of a lane, a direction loss is incorporated, which measures the cosine similarity. The loss is formulated as

\begin{equation}
    \mathcal{L}_{\text{dir}} = \sum_{i}^{N} \mathds{1}_{\{ c_i \neq \varnothing \}} \sum_{j}^{N_v} \text{cos\_sim} (\hat{y}_{\hat{\sigma}_{(i)},j}, y_{i, \sigma_{(j)}}),
\end{equation}

\noindent where the cosine similarity is calculated based on the direction vector of two consecutive points over the centerline and lane dividers. The resulting loss $L_{\text{o2o}}$, based on the computed one-to-one assignment $\hat{\sigma}$, is defined as

\begin{equation}
    \mathcal{L}_{\text{o2o}} = \lambda_{\text{class}} \mathcal{L}_{\text{class}} + \lambda_{\text{point}} \mathcal{L}_{\text{point}} + \lambda_{\text{dir}} \mathcal{L}_{\text{dir}},
\end{equation}

\noindent where $\lambda_{\text{class}}, \lambda_{\text{point}}, \text{and } \lambda_{\text{dir}}$ are balancing weights for the distinct loss terms. To accelerate convergence, we add a one-to-many auxiliary loss $\mathcal{L}_{\text{o2m}}$ that builds upon $\mathcal{L}_{\text{o2o}}$, but allows to reuse the same ground truth multiple times, as described in \cite{maptrv2}. Furthermore, a second auxiliary loss $\mathcal{L}_{\text{aux}}$ is added that acts as $\mathcal{L}_{\text{o2o}}$, but is calculated from the results of the intermediate transformer decoder hidden states instead of only the last hidden state. Consequently, the final training loss is defined as

\begin{equation}
    \mathcal{L} = \lambda_{\text{o2o}} \mathcal{L}_{\text{o2o}} + \lambda_{\text{o2m}} \mathcal{L}_{\text{o2m}} + \lambda_{\text{aux}} \mathcal{L}_{\text{aux}},
\end{equation}

\noindent where $\lambda_{\text{o2o}}, \lambda_{\text{o2m}}, \text{and } \lambda_{\text{aux}}$ are balancing weights for the individual loss terms.

\subsection{Implementation}
All tiles are extracted within a range of $[-30\textit{m}, +30\textit{m}]$ along both x- and y-directions. Ground truth centerlines and lane dividers are resampled to $M = 20$ points for training and evaluation. The input image resolution is $512 \times 512$. We use a ResNet-50 backbone unless otherwise specified. The transformer embedding dimension is 256. For deformable attention, we use 4 sampling points and 8 attention heads. Both encoder and decoder consist of 6 layers. All feed-forward networks use an embedding dimension of 512 with a dropout rate of 0.1 applied throughout the transformer. Training is performed with batch size 32 using AdamW and an initial learning rate of $1 \times 10^{-4}$ with cosine decay. The backbone learning rate is set to one tenth of this value. The number of queries is $N = 50$ for the one-to-one group and 150 for the one-to-many group. All loss weights $\lambda_{\text{class}}, \lambda_{\text{point}}, \lambda_{\text{dir}}, \lambda_{\text{o2o}}, \lambda_{\text{o2m}}, \lambda_{\text{aux}}$ are set to 1.

\subsection{Metric}
Following common practice in HD map generation, we evaluate the predictions using average precision based on the Chamfer distance under multiple thresholds \cite{tang_map_survey}. The average precision is defined as

\begin{equation}
    AP = \frac{1}{|T|} \sum_{\tau \in T} AP_\tau,
\end{equation}

\noindent where $T$ contains the thresholds $\{0.5, 1.0, 1.5\}$. We evaluate the average precision separately for centerlines and lane dividers, denoted as $AP_c$ and $AP_{ld}$, respectively.

\subsection{Quantitative Results}

\begin{table}[!b]
    \caption{Training results on different datasets.}
    \centering
    \fontsize{8.75}{12}\selectfont
    \begin{tabularx}{\linewidth}{l|X|C|C|C}
        \toprule
        \textbf{Train Split} & \textbf{Val Split} & $\textbf{AP}_c \uparrow$ & $\textbf{AP}_{ld}\uparrow$ & $\textbf{AP}\uparrow$ \\
        \midrule
        \midrule
        Internal & Internal & 81.3 & 76.9 & 79.1 \\
        nuPlan & nuPlan & 44.7 & 43.7 & 44.2 \\
        \midrule
        nuScenes & nuScenes & 64.2 & \textbf{60.8} & 62.5 \\
        nuScenes $+$ Internal & nuScenes & \textbf{65.5} & 60.4 & \textbf{63.0} \\
        nuScenes $+$ nuPlan & nuScenes & 61.8 & 58.5 & 60.2 \\
        \bottomrule
    \end{tabularx}
    \label{quant_results_main}
\end{table}

Training is conducted on individual datasets and their combinations to study the effect of cross-dataset training on validation performance. \autoref{quant_results_main} reports the AP for different training and validation splits. The relatively high AP of 79.1 on the internal dataset is mainly due to simple scenes with straight lanes (e.g., highways) and tiles containing fewer lanes to predict. In contrast, nuPlan achieves an AP of 44.2, featuring a significantly higher trajectory density and, consequently, more complex road structures with a larger number of lanes per tile. The nuScenes dataset yields an AP of 62.5, which can also be attributed to simpler sampled tiles and a lower average number of lanes per tile. Furthermore, we investigate the influence on the nuScenes validation split when extending its training split with training data from other datasets. In \autoref{quant_results_main}, it can be seen that incorporating the internal dataset improves the overall AP. However, when considering lane dividers specifically, a slight decrease can be observed. While the $\text{AP}_\text{c}$ improves, potentially due to increased data diversity, this benefit does not fully transfer to $\text{AP}_\text{ld}$. A possible reason is the variation in road widths across datasets, which may limit the model’s ability to generalize accurate lane divider positions. In contrast, when extending the nuScenes training data with that of nuPlan, the AP decreases to 60.2, which can possibly be attributed to the different data distributions in terms of number of vehicle trajectories and number of lanes to predict per tile.

\subsection{Qualitative Results}
\begin{figure*}[!b]
    \centering
    \includegraphics[width=\textwidth]{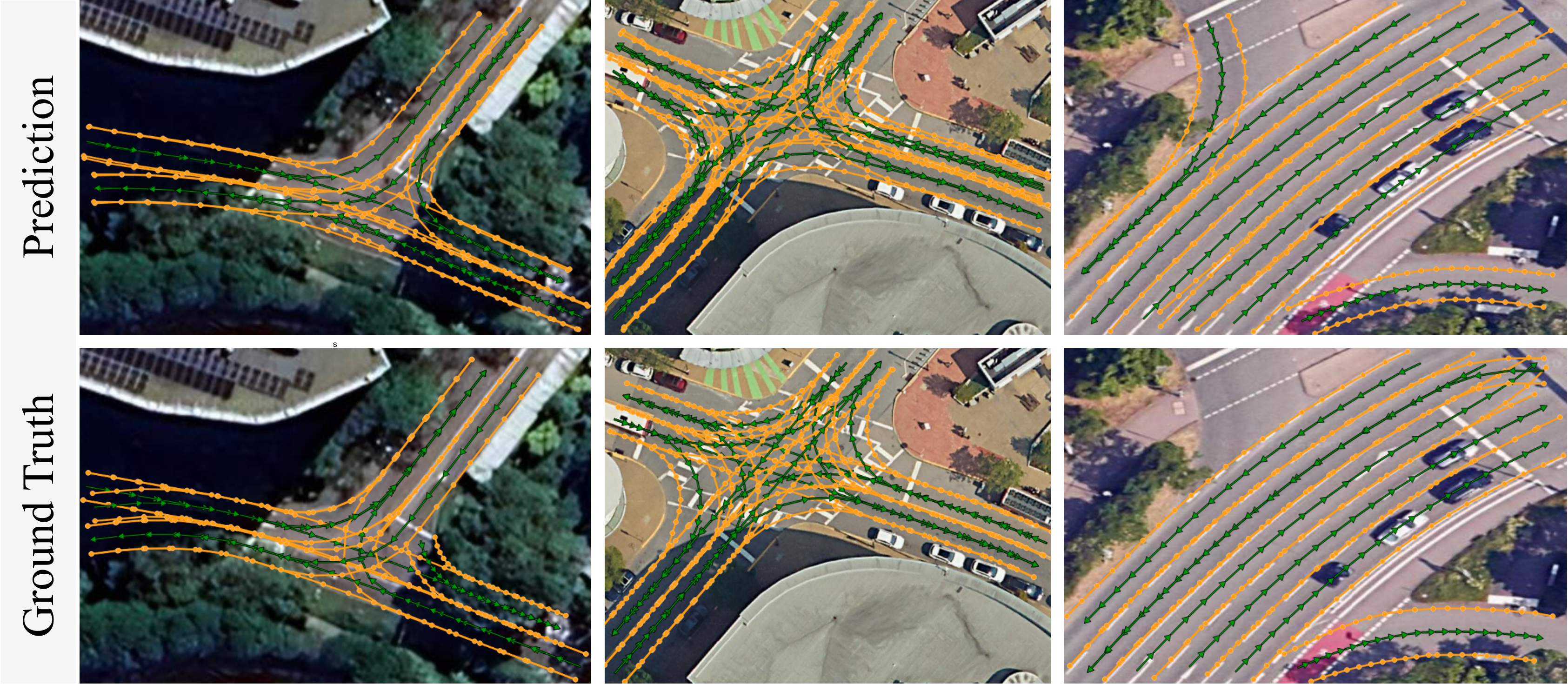} 
    \caption{Examples of model outputs for selected tiles. The left and middle columns originate from nuScenes and nuPlan, respectively, while the right column is from the internal dataset. Lanes are predicted with direction, consisting of centerlines (green) and lane dividers (orange). Lanes with the highest confidence scores are depicted.}
    \label{fig:results}
\end{figure*}

\autoref{fig:results} shows examples of model outputs. Green lines represent centerlines, while orange lines represent lane dividers. Each line is directed. We sort the predicted lanes by confidence scores from the classification branch and visualize only those with the highest scores. The model correctly captures the overall lane geometry and direction. However, in the left column, one lane is missing compared to the ground truth, indicating room for improvement. Another limitation is the lack of geometric consistency in merging centerlines and lane dividers, where lines that should merge are slightly misaligned or remain separated. This could be addressed by explicitly supervising such overlapping regions during the merging process.

\subsection{Ablation Studies}
For the ablation studies, we use the near-extrapolation split of nuScenes for training. \autoref{quant_results_backbone} reports the AP results for different backbone networks used for feature extraction. For our setup, ResNet-50 achieves the best performance with an AP of 62.5. Using a lighter backbone such as ResNet-34 results in an AP of 49.7, representing a significant drop compared to ResNet-50. Swin-T and EfficientNet-B5 achieve APs of 47.7 and 40.4, respectively. These results indicate that ResNet-50 is the most suitable backbone for our task. 


\begin{table}[t!]
    \caption{Results using different backbones. All training runs conducted on nuScenes dataset using the near extrapolation split.}
    \centering
    \fontsize{8.75}{12}\selectfont
    \begin{tabularx}{\linewidth}{X|C|C|C}
        \toprule
        \textbf{Backbone} & $\textbf{AP}_c \uparrow$ & $\textbf{AP}_{ld}\uparrow$ & $\textbf{AP}\uparrow$ \\
        \midrule
        \midrule
        ResNet-34 & 51.1 & 48.3 & 49.7 \\
        ResNet-50 & \textbf{64.2} & \textbf{60.8} & \textbf{62.5} \\
        SwinT & 49.1 & 46.2 & 47.7 \\
        EfficientNet-B5 & 41.1 & 39.7 & 40.4 \\
        \bottomrule
    \end{tabularx}
    \label{quant_results_backbone}
\end{table}

Further ablation studies are shown in \autoref{different_methods}, using training on the near-extrapolation split of nuScenes with data augmentation as the baseline. Removing data augmentation decreases AP by 11.8. We also evaluate query pruning, following \cite{xu_gpq_2025}, which removes low-confidence queries during training. Training with query pruning was conducted with data augmentation, and pruning was applied only to the one-to-one group. The number of queries was reduced from 50 to 25, resulting in an AP drop of 3.7.

\begin{table}[t!]
    \caption{Results using different methods. All training runs conducted on nuScenes dataset using the near extrapolation split.}
    \centering
    \fontsize{8.75}{12}\selectfont
    \begin{tabularx}{\linewidth}{X|C|C|C}
        \toprule
        \textbf{Method} & $\textbf{AP}_c \uparrow$ & $\textbf{AP}_{ld}\uparrow$ & $\textbf{AP}\uparrow$ \\
        \midrule
        \midrule
        With Data Aug. & 64.2 & 60.8 & 62.5 \\
        Without Data Aug. & 52.5 & 48.9 & 50.7  \\
        Query Pruning & 60.0  & 57.6 & 58.8 \\
        \bottomrule
    \end{tabularx}
    \label{different_methods}
\end{table}


%% file: content/conclusion.tex
\section{CONCLUSION}
This work introduced a pipeline for tile extraction from a global HD map, where each tile was transformed into a rasterized representation encoding both the presence and directional information of trajectories from crowdsourced datasets. The model predicted lane representations consisting of a centerline and corresponding left and right lane dividers, with dividers derived from predicted offsets. Experiments on multiple datasets with varying trajectory densities and numbers of ground truth lanes demonstrated the impact of data characteristics on performance. Future work could explore additional data sources such as satellite imagery or OpenStreetMap (OSM) to improve accuracy, particularly for lane divider localization, and extend the output set to include road boundaries and pedestrian crossings. Leveraging OSM for tile selection could further focus on complex structures and improve performance on challenging scenarios.

%% file: content/acknowledgment.tex
\section*{Acknowledgment}
This work is a result of the joint research project STADT:up (19A22006O). The project is supported by the German Federal Ministry for Economic Affairs and Energy (BMWE), based on a decision of the German Bundestag. The author is solely responsible for the content of this publication.

%% file: bib.bib
@inproceedings{Hao_25,
    author={Hao, Xiaoshuai and Zhao, Yuting and Ji, Yuheng and Dai, Luanyuan and Hao, Peng and Li, Dingzhe and Cheng, Shuai and Yin, Rong},
    booktitle={2025 IEEE/RSJ International Conference on Intelligent Robots and Systems (IROS)}, 
    title={What Really Matters for Robust Multi-Sensor HD Map Construction?}, 
    year={2025},
    volume={},
    number={},
    pages={1298-1304}
}

@inproceedings{lanesegnet,
    title={LaneSegNet: Map Learning with Lane Segment Perception for Autonomous Driving},
    author={Li, Tianyu and Jia, Peijin and Wang, Bangjun and Chen, Li and Jiang, Kun and Yan, Junchi and Li, Hongyang},
    booktitle={The Fourteenth International Conference on Learning Representations (ICLR)},
    year={2024}
}

@inproceedings{lane_gap,
    author="Liao, Bencheng
    and Chen, Shaoyu
    and Jiang, Bo
    and Cheng, Tianheng
    and Zhang, Qian
    and Liu, Wenyu
    and Huang, Chang
    and Wang, Xinggang",
    title="Lane Graph as Path: Continuity-Preserving Path-Wise Modeling for Online Lane Graph Construction",
    booktitle="Computer Vision -- ECCV 2024",
    year="2025",
    publisher="Springer Nature Switzerland",
    address="Cham",
    pages="334--351",
    isbn="978-3-031-72784-9"
}

@inproceedings{TrailTR,
    author={Hubbertz, Michael and Colling, Pascal and Han, Qi and Meisen, Tobias},
    title={Inferring Driving Maps by Deep Learning-based Trail Map Extraction},
    booktitle={Proceedings of the IEEE/CVF Conference on Computer Vision and Pattern Recognition (CVPR) Workshops},
    month={June},
    year={2025},
    pages={2450-2459}
}

@inproceedings{lmtnet,
    author={Mink, Michael and Monninger, Thomas and Staab, Steffen},
    booktitle={2024 IEEE 27th International Conference on Intelligent Transportation Systems (ITSC)}, 
    title={LMT-Net: Lane Model Transformer Network for Automated HD Mapping from Sparse Vehicle Observations}, 
    year={2024},
    volume={},
    number={},
    pages={1203-1210}
}

@inproceedings{detr,
    author="Carion, Nicolas
    and Massa, Francisco
    and Synnaeve, Gabriel
    and Usunier, Nicolas
    and Kirillov, Alexander
    and Zagoruyko, Sergey",
    title="End-to-End Object Detection with Transformers",
    booktitle="Computer Vision -- ECCV 2020",
    year="2020",
    publisher="Springer International Publishing",
    address="Cham",
    pages="213--229",
    isbn="978-3-030-58452-8"
}

@article{local_hd_gen_survey,
    author={Zhang, Yangrong and Qian, Yeqiang and Meng, Chao and Zhang, Rui and Yi, Hongjun and Wang, Chunxiang and Yang, Ming},
    journal={IEEE Transactions on Intelligent Transportation Systems}, 
    title={Local Vectorized High Definition Map Construction for Autonomous Driving: A Comprehensive Review}, 
    year={2025},
    volume={26},
    number={12},
    pages={21502-21525}
}

@article{Qin_23,
    author={Qin, Tong and Huang, Haihui and Wang, Ziqiang and Chen, Tongqing and Ding, Wenchao},
    journal={IEEE Robotics and Automation Letters}, 
    title={Traffic Flow-Based Crowdsourced Mapping in Complex Urban Scenario}, 
    year={2023},
    volume={8},
    number={8},
    pages={5077-5083}
}

@inproceedings{vectormapnet,
    title={VectorMapNet: End-to-end Vectorized HD Map Learning},
    author={Liu, Yicheng and Yuan, Tianyuan and Wang, Yue and Wang, Yilun and Zhao, Hang},
    booktitle={International conference on machine learning},
    year={2023},
    organization={PMLR}
}

@inproceedings{maptrv,
    title={MapTR: Structured Modeling and Learning for Online Vectorized HD Map Construction},
    author={Liao, Bencheng and Chen, Shaoyu and Wang, Xinggang and Cheng, Tianheng and Zhang, Qian and Liu, Wenyu and Huang, Chang},
    booktitle={International Conference on Learning Representations},
    year={2023}
}

@inproceedings{graph_as_path,
    title={Lane Graph as Path: Continuity-preserving Path-wise Modeling for Online Lane Graph Construction},
    author={Bencheng Liao and Shaoyu Chen and Bo Jiang and Tianheng Cheng and Qian Zhang and Wenyu Liu and Chang Huang and Xinggang Wang},
    booktitle={European Conference on Computer Vision},
    year={2023}
}

@inproceedings{streammapnet,
    author={Yuan, Tianyuan and Liu, Yicheng and Wang, Yue and Wang, Yilun and Zhao, Hang},
    title={StreamMapNet: Streaming Mapping Network for Vectorized Online HD Map Construction},
    booktitle={Proceedings of the IEEE/CVF Winter Conference on Applications of Computer Vision (WACV)},
    month={January},
    year={2024},
    pages={7356-7365}
}

@article{maptrv2,
    title={Maptrv2: An end-to-end framework for online vectorized hd map construction},
    author={Liao, Bencheng and Chen, Shaoyu and Zhang, Yunchi and Jiang, Bo and Zhang, Qian and Liu, Wenyu and Huang, Chang and Wang, Xinggang},
    journal={International Journal of Computer Vision},
    pages={1--23},
    year={2024},
    publisher={Springer}
}

@article{mapcvv,
    author={Chen, Pengxin and Jiang, Xiaoqi and Zhang, Yingjun and Tan, Jiahao and Jiang, Rong},
    journal={IEEE Robotics and Automation Letters}, 
    title={MapCVV: On-Cloud Map Construction Using Crowdsourcing Visual Vectorized Elements Towards Autonomous Driving}, 
    year={2024},
    volume={9},
    number={6},
    pages={5735-5742}
}

@article{elghazaly_map_survey,
    author={Elghazaly, Gamal and Frank, Raphaël and Harvey, Scott and Safko, Stefan},
    journal={IEEE Open Journal of Intelligent Transportation Systems}, 
    title={High-Definition Maps: Comprehensive Survey, Challenges, and Future Perspectives}, 
    year={2023},
    volume={4},
    number={},
    pages={527-550}
}

@article{lyu_map_generation_survey,
    AUTHOR = {Lyu, Hongyu and Berrio Perez, Julie Stephany and Huang, Yaoqi and Li, Kunming and Shan, Mao and Worrall, Stewart},
    TITLE = {Online High-Definition Map Construction for Autonomous Vehicles: A Comprehensive Survey},
    JOURNAL = {Journal of Sensor and Actuator Networks},
    VOLUME = {14},
    YEAR = {2025},
    NUMBER = {1},
    ARTICLE-NUMBER = {15},
    ISSN = {2224-2708}
}

@article{guo_map_generation_survey,
    AUTHOR = {Guo, Yuan and Zhou, Jian and Li, Xicheng and Tang, Youchen and Lv, Zhicheng},
    TITLE = {A Review of Crowdsourcing Update Methods for High-Definition Maps},
    JOURNAL = {ISPRS International Journal of Geo-Information},
    VOLUME = {13},
    YEAR = {2024},
    NUMBER = {3},
    ARTICLE-NUMBER = {104},
    ISSN = {2220-9964}
}

@article{tang_map_survey,
    author={Tang, Xuewei and Jiang, Kun and Yang, Mengmeng and Liu, Zhaoyang and Jia, Peijin and Wijaya, Benny and Wen, Tuopu and Cui, Le and Yang, Diange},
    journal={IEEE Transactions on Intelligent Vehicles}, 
    title={High-Definition Maps Construction Based on Visual Sensor: A Comprehensive Survey}, 
    year={2024},
    volume={9},
    number={10},
    pages={5973-5994}
}

@inproceedings{nuscenes_dataset_split,
    author    = {Lilja, Adam and Fu, Junsheng and Stenborg, Erik and Hammarstrand, Lars},
    title     = {Localization Is All You Evaluate: Data Leakage in Online Mapping Datasets and How to Fix It},
    booktitle = {Proceedings of the IEEE/CVF Conference on Computer Vision and Pattern Recognition (CVPR)},
    month     = {June},
    year      = {2024},
    pages     = {22150-22159}
}

@inproceedings{djuric_2020,
  author={Djuric, Nemanja and Radosavljevic, Vladan and Cui, Henggang and Nguyen, Thi and Chou, Fang-Chieh and Lin, Tsung-Han and Singh, Nitin and Schneider, Jeff},
  booktitle={2020 IEEE Winter Conference on Applications of Computer Vision (WACV)}, 
  title={Uncertainty-aware Short-term Motion Prediction of Traffic Actors for Autonomous Driving}, 
  year={2020},
  volume={},
  number={},
  pages={2084-2093}
}

@inproceedings{fritz_advntg_2025,
	address = {Gold Coast, Australia},
	title = {{ADVNTG}: {Autonomous} {Driving} {Vehicle} and {Neural} {Transformer}-{Based} {HD} {Map} {Generation} {Using} {Crowd}-{Sourced} {Fleet} {Data}},
	copyright = {https://doi.org/10.15223/policy-029},
	isbn = {979-8-3315-2418-0},
	shorttitle = {{ADVNTG}},
	urldate = {2026-04-09},
	booktitle = {2025 {IEEE} 28th {International} {Conference} on {Intelligent} {Transportation} {Systems} ({ITSC})},
	publisher = {IEEE},
	author = {Fritz, Daniel and Lagamtzis, Dimitrios and Mink, Michael and Schober, Steffen},
	month = nov,
	year = {2025},
	pages = {1954--1959},
}

@article{bai_2026,
  author={Bai, Wenjing and Zhang, Yunzhou and Guo, Qi and Liu, Wei and Du, Shangwei and Hu, Jun and Cheng, Shuai and Ning, Zuotao},
  journal={IEEE Transactions on Multimedia}, 
  title={Dynamic Query Management and Internal Consistency Representation based Transformer for Online Vectorized HD Map Construction}, 
  year={2026},
  volume={},
  number={},
  pages={1-15},
}

@article{Lie_2026,
  author={Liu, Guangwei and Zhang, Dazhi and Xu, Chengjian and Zhang, Xiaoyu and Zhang, Zichao and Zhao, Ji and Wu, Zheng and Zhang, Jian},
  journal={IEEE Robotics and Automation Letters}, 
  title={City-Scale Lane-Level Mapping From Crowdsourced Trajectories and Satellite Imagery}, 
  year={2026},
  volume={11},
  number={4},
  pages={4793-4800}
}

@inproceedings{xu_gpq_2025,
    title={Redundant Queries in DETR-Based 3D Detection Methods: Unnecessary and Prunable}, 
    author={Lizhen Xu and Zehao Wu and Wenzhao Qiu and Shanmin Pang and Xiuxiu Bai and Kuizhi Mei and Jianru Xue},
    booktitle={Proceedings of the AAAI Conference on Artificial Intelligence},
    year={2025},
}
